\documentclass[letterpaper, 10 pt, conference]{ieeeconf}  

\IEEEoverridecommandlockouts                              

\usepackage{graphics} 
\usepackage{epsfig} 
\usepackage{mathptmx} 
\usepackage{times} 
\usepackage{amsmath} 
\usepackage{amssymb}  
\usepackage{caption} 
\usepackage{subcaption}
\usepackage{xcolor}
\title{\LARGE \bf
The Effects of Robot Motion on Comfort Dynamics of Novice Users in Close-Proximity Human-Robot Interaction
}

\author{Pierce Howell, Jack Kolb$^{*}$,  Yifan Liu$^{*}$, and Harish Ravichandar
\thanks{*These authors contributed equally to this work.}
\thanks{All authors are with Georgia Institute of Technology, USA. Correspondence: {\tt pierce.howell@gatech.edu}}%
}

\begin{document}

\maketitle
\thispagestyle{empty}
\pagestyle{empty}

\begin{abstract}
Effective and fluent close-proximity human-robot interaction requires understanding how humans get habituated to robots and how robot motion affects human comfort.  While prior work has identified humans' preferences over robot motion characteristics and studied their influence on comfort, we are yet to understand how novice \textit{first-time} robot users get habituated to robots and how robot motion impacts the \textit{dynamics} of comfort over repeated interactions. To take the first step towards such understanding, we carry out a user study to investigate the connections between robot motion and user comfort and habituation. Specifically, we study the influence of workspace overlap, end-effector speed, and robot motion legibility on overall comfort and its evolution over repeated interactions. Our analyses reveal that workspace overlap, in contrast to speed and legibility, has a significant impact on users' perceived comfort and habituation. In particular, lower workspace overlap leads to users reporting significantly higher overall comfort, lower variations in comfort, and fewer fluctuations in comfort levels during habituation.
\end{abstract}

\section{Introduction} \label{Introduction}

As robots increasingly become an integral part of our personal and professional spaces, it is essential to study how their presence impacts people's sense of comfort. This need is particularly important in environments involving close-proximity interactions. Indeed, an improved understanding of how robots influence user comfort can help us design better collaborative robots and improve the fluency and effectiveness of interactions.

Several prior efforts have studied the impact of robot motion characteristics on user comfort (see Sec. \ref{sec:related_work} for a detailed discussion). Motion characteristics that have been studied include human-to-robot distance~\cite{araiAssessmentOperatorStress2010}, robot speed~\cite{zoghbiEvaluationAffectiveState2009}, motion fluency~\cite{draganEffectsRobotMotion2015}, and direction of approach \cite{zoghbiEvaluationAffectiveState2009}. However, most existing studies have two important limitations. First, they tend to assume comfort to be a static quantity and do not study how comfort evolves over repeated interactions. Second, they do not exclusively focus on novice users who have no prior experience working with a large physical robot. As such, we lack an understanding of how novice users get habituated to robots, and how this habituation is impacted by the robot's motion characteristics. Such understanding will help roboticists design robot behaviors such that first-time users develop a positive attitude towards collaborative robots and improve the chances of adoption.

In this paper, we report our findings from a 40-subject user study on the impact of robot motion characteristics on \emph{first-time} robot users' perceived comfort and habituation when engaged in repetitive close-proximity interactions.  We employed a between-subjects design because we wish to capture how users get habituated to robots as opposed to identifying user preferences over motion characteristics.

We specifically study the impact of \emph{workspace overlap}, \emph{robot speed}, and \emph{robot motion legibility} (the independent variables) on users' perceived comfort (the dependent variable). To this end, we defined two levels of each independent variable (low vs high workspace overlap, nominal vs. fast robot speed, and predictable vs. legible robot motion). We measured user comfort in a mixed response format involving visual-analog scales, Likert scales, pre- and post-experiment questionnaires, and behavior analyses. We repeatedly measure comfort after every episode of interaction and analyze how motion characteristics influence overall comfort and changes in comfort across episodes.

Rigorous statistical analyses reveal that varying the amount of workspace overlap had the most pronounced impact on the user's subjective comfort. Specifically, we find that higher workspace overlap results in significantly lower overall comfort, and more frequent fluctuations in comfort levels over repeated interactions. Surprisingly, we did not find sufficient evidence to believe that robot speed and motion legibility impact user comfort and habituation. 
We believe that these findings can help inform how adaptation algorithms and default robot behaviors should be designed to effectively introduce robots to novice users.

\begin{figure}[t]
    \centering
    \includegraphics[width=0.8\linewidth]{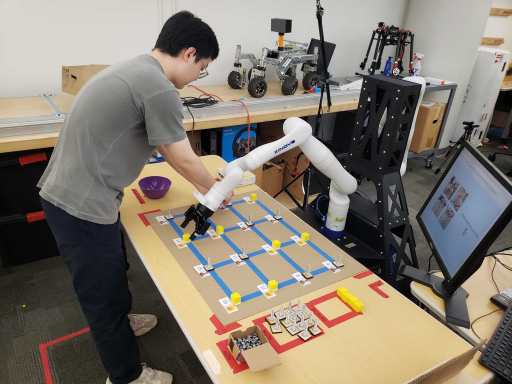}
    \caption{We study the influence of robot motion characteristics on the perceived comfort and habituation dynamics of novice users who have never interacted with large physical robots before.}
    \label{fig:working-with-robot}
\end{figure}

\section{Related Work} \label{sec:related_work}

In this section, we contextualize our contributions within different relevant categories of prior work.

\vspace{2pt}
\noindent \textbf{\textit{Effect of motion on comfort}}: 
Prior work has demonstrated that robot speed~\cite{zoghbiEvaluationAffectiveState2009}, proximity~\cite{araiAssessmentOperatorStress2010}, and legibility~\cite{draganEffectsRobotMotion2015} exert significant influence on users' subjective comfort. The degree to which these characteristics affect users' comfort depends on a variety of factors (e.g. prior robot experience, robot type, task context, characteristic joint relationships). For example, studies on comfort and robot acceptance show that the combined relationship between robot speed and distance is strongly impactful, where decreased speeds at shorter distances result in increased levels of comfort and acceptance~\cite{araiAssessmentOperatorStress2010, brandlHumanRobotInteractionAssisted2016}. However, existing studies traditionally treat comfort as a static quantity.
In contrast, we study how comfort evolves over repeated interactions and the effect of three motion characteristics (workspace overlap, speed, and legibility) on this evolution.
Further, most studies utilized a within-participant design and focused on capturing user preferences over motion characteristics. In contrast, we present a between-participant study and focus on capturing users' habituation dynamics.

\vspace{2pt}
\noindent \textbf{\textit{Modeling trust}}:
Prior work has studied the dynamics of user trust in autonomous systems to develop personalized trust prediction models~\cite{leeTrustControlStrategies1992, xuOPTIMoOnlineProbabilistic2015, guoModelingPredictingTrust2021}, which have proven valuable in developing adaptive autonomous systems that improve trust, performance, and user experience~\cite{chenPlanningTrustHumanRobot2018}. 
These works show that trust evolves as a function of i) prior trust, ii) task performance or reliability, and iii) automation failures~\cite{yangEvaluatingEffectsUser2017a}. Yet, trust and comfort are two distinctly different concepts of user experience~\cite{pineauPsychologicalMeaningComfort1982}. Trust defines a user's reliance, confidence, and willingness to be vulnerable, whereas a user's comfort is defined as their physiological sense of well-being and state of ease.
In contrast to prior work characterizing the evolution of trust, we study how user comfort evolves over repeated interaction as a function of robot motion characteristics in close-proximity.  Additionally, prior work in trust dynamics considers the human in a supervisory role over the autonomous system.
In our study, the user and robot interact as collaborators, wherein the user does not have the ability to command or control the robot.

\vspace{2pt}
\noindent \textbf{\textit{Measurement of comfort}}:
There are four traditional methods to measure comfort in human-robot interactions: i) questionnaires, ii) reporting devices, iii) behavioral analysis, and iv) physiological measurements, and each has unique benefits and limitations. Comfort questionnaires (e.g., \emph{GODSPEED}~\cite{bartneckMeasurementInstrumentsAnthropomorphism2009a} and \emph{Robotic Social Attributes Scale (RoSAS)}~\cite{carpinellaRoboticSocialAttributes2017})
use multi-item scales and are administered offline to probe users pre- and post-condition. However, such questionnaires may not be well suited to repeatedly measure comfort due to their high cognitive demand~\cite{friborgLikertbasedVsSemantic2006}. 
To mitigate this problem, researchers have used physical hand-held devices that enable repeated measures of comfort and affective state. Examples include a hand-held comfort level device using single slider labeled with a sad face and a happy face on either end, enabling users to continually report their comfort while holding the device~\cite{koayEmpiricalResultsUsing2006}, and a handheld joystick that measured users' affect (valence and arousal) in close-proximity observation of a robot arm~\cite{zoghbiEvaluationAffectiveState2009}.
However, these devices are not appropriate for close-proximity collaboration since a human must use both hands and pay attention to the task. 
Alternatively, behavioral analysis offers a non-intrusive way (e.g.,video analysis) to estimate comfort. User proximity~\cite{mummHumanrobotProxemicsPhysical2011}, idle time~\cite{lasotaAnalyzingEffectsHumanAware2015}, and body language~\cite{koayEmpiricalResultsUsing2006} are but a few examples. While some of these measures likely correlate with users' comfort, they might not reliably capture how a user feels.
Finally, physiological measures (e.g., heart rate variability and galvanic skin response) attempt to measure emotional states which are less susceptible to reporting biases~\cite{bethelSurveyPsychophysiologyMeasurements2007}. 
However, estimating emotions via physiological measures continues to be challenging; comfort inference from such measures often do not align with subjective reports~\cite{kulicPhysiologicalSubjectiveResponses2007}, and wearing the necessary sensors can be cumbersome.
In our work, we use a combination of repeated comfort feedback, behavioral measures, and multi-item questionnaires to study the comfort and habituation dynamics of novice users in close-proximity settings. 


\section{Research Questions} \label{Research-Questions}

We are interested in exploring how motion characteristics of a robot manipulator affect how novice users become habituated to and feel comfortable working with robots in close-proximity. We are particularly interested in scenarios in which the workspaces of the robot arm and human overlap considerably. We focus on three motion characteristics: robot speed, workspace overlap, and legibility.

First, we aim to gain insight into how comfortable users feel as they start working with a robot. Our population of interest is non-expert, novice users who have little to no prior experience working with robots. To this end, we pose our first research question:

\noindent \textbf{RQ1}: How do speed, workspace overlap, and legibility influence \textit{first-time} robot users' perceived comfort? Do the motion characteristics exert their influence independently, or are there interaction effects?

We also investigate the dynamics of comfort by studying how users' comfort levels evolve as they engage in repeated interactions. We pose our second research question:

\noindent \textbf{RQ2}: How do speed, workspace overlap, and legibility affect the \textit{evolution} of perceived comfort in novice users over repeated interactions?

\section{Study Design} \label{Study-Design}

\begin{figure}[t]
    \centering
    \includegraphics[width=0.8\linewidth]{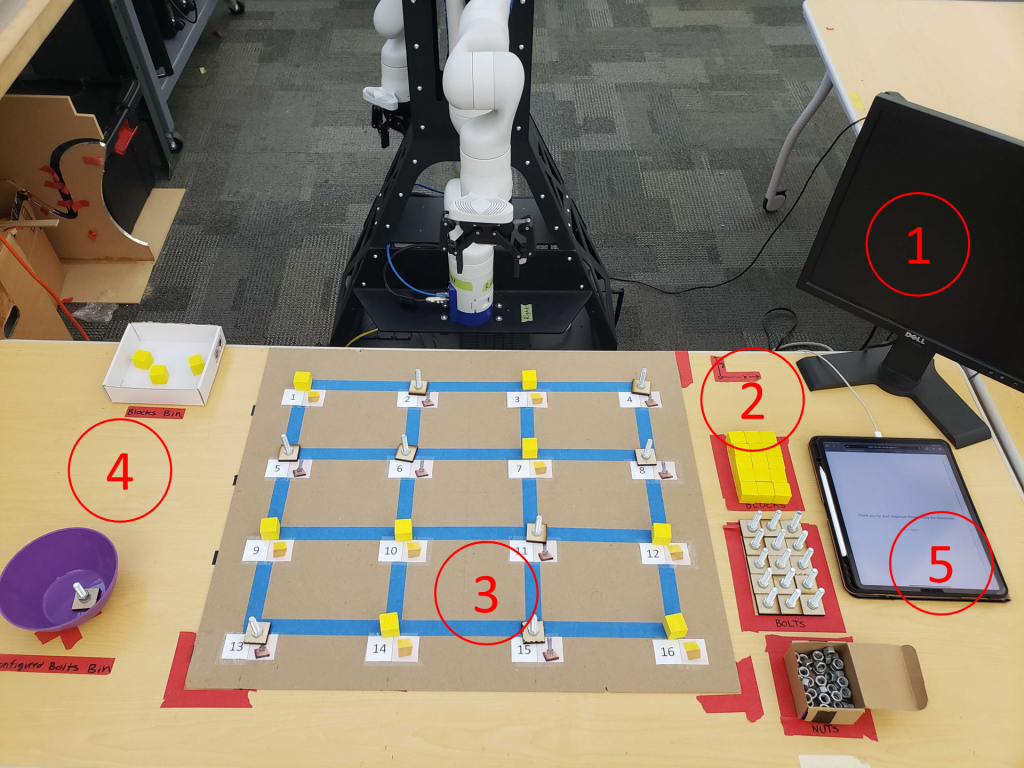}
    \caption{Experimental setup from the subjects' perspective includes: 1) display to identify the bolt to be configured, 2) pickup area for nuts, bolts, and blocks, 3) shared manipulation workspace, 4) bins to deposit configured items, and 5) a tablet for the questionnaire.}
    \label{fig:experiment-setup}
\end{figure}

We conducted a between-subjects study involving first-time robot users working with a 7 degree-of-freedom Kinova Gen3 robot arm to complete an assembly and binning task in a close-proximity setting with a shared workspace (see Fig. \ref{fig:experiment-setup}). 

\subsection{Task} \label{Task}

We designed the task and environment to elicit behaviors that would generally arise in close-proximity interactions with overlapping human-robot workspaces, such as those encountered in assembly or manufacturing settings. For the task, subjects were required to screw nuts onto specified bolts on a tabletop surface (termed ``configuring"), while the robot simultaneously removed blocks from the workspace. The robot was positioned opposite the table and faced the user, with the robot's base approximately 1 meter from the subject's torso. Fig. \ref{fig:experiment-setup} shows the workspace from the subject's point of view. During each interaction, participants perform the following tasks sequentially. 
\begin{enumerate}
    \item Configure a bolt with a nut without moving the bolt (i.e. keep the bolt in place while configuring).
    \item Place the configured bolt in a bin and replace it with a new unconfigured bolt.
    \item Place a new yellow block in the location from which the robot moved the previous block.
\end{enumerate}
Concurrently, the robot performs the following tasks.
\begin{enumerate}
    \item Pick up one of the yellow blocks from an unspecified location of the grid.
    \item Place the block in the robot block deposit bin.
    \item Move back to the robot's home position.
\end{enumerate}

We term an ``episode" as one round of the subject configuring and removing a specified bolt and the robot removing one block. Each subject completed 12 episodes and answered a brief questionnaire after each episode. 

Each position on the workspace was labeled numerically, and a screen in front of the user displayed the position of the bolt to configure. An audio cue signalled the start of each episode. To study the effect of robot trajectory legibility, the user was uninformed which block the robot would remove. After the subject completed configuring their bolt, they would remove the bolt and replace both the bolt and the block from a supply area.

\subsection{Experiment Conditions} \label{Experiment-Conditions}

We designed three independent variables (IVs):  i) end-effector speed, ii) workspace overlap, and iii) legibility.

\vspace{2pt}
\noindent \emph{End-effector speed}: we designed a \texttt{normal (N)} $(M=98.13,\ SD=0.9596)$ and \texttt{fast (F)} $(M=191.4056, \ SD=16.1912)$ level for speed, both measured in $mm/s$ and represent the average magnitude of end-effector Cartesian velocity. 
To ensure safety, we kept the maximum end-effector speed under 500 mm/s as recommended by~\cite{kochApproachAutomatedSafety2019}.

\begin{figure}
    \centering
    \includegraphics[width=0.8\linewidth]{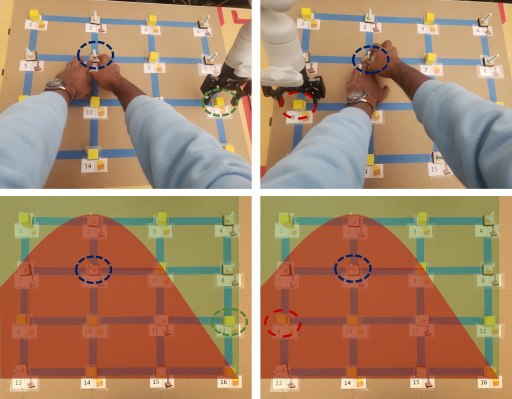}
    \caption{Low (left) and high (right) workspace overlap depending on whether the robot's target location is inside or outside of the user's workspace (red).
    }
    \label{fig:workspace-proximity}
\end{figure}

\vspace{2pt}
\noindent \emph{Workspace overlap}: we defined \texttt{low (L)} and a \texttt{high (H)} overlap conditions based on the expected intersection of the workspace of the robot end-effector and the human hand. 
Specifically, we defined the levels based on regions defined by a parabolic-like curve traced by the subject's hands when configuring the bolt as shown in Fig. \ref{fig:workspace-proximity}. The red area in Fig. \ref{fig:workspace-proximity} is considered be high overlap since operating in this region is likely to result in close and intersecting movements.

\vspace{2pt}
\noindent \emph{Trajectory legibility}: we designed \texttt{predictable (P)} and \texttt{legible (L)} categories based upon the definitions from \cite{draganEffectsRobotMotion2015}. 
We designed predictable motion as straight-line end-effector paths from the home position to the target block, as that's the trajectory a user would expect if they knew the target location. To design legible motions that would help users predict the target location, we used the grid structure of our setup. We designed legible trajectories in which the robot first moved to line up with the column of the target block, at a height proportional to its distance to the target block down the column. We believe such motions are legible given that the block-bolt grid on the workspace is a highly-salient environmental feature, and thus aligning the robot' pick trajectory to this grid makes the robot's motion and intention predictable (i.e. more legible).

\subsection{Subjective Measures} \label{Subjective Measures}

\begin{figure}
    \centering
    \includegraphics[width=0.8\columnwidth]{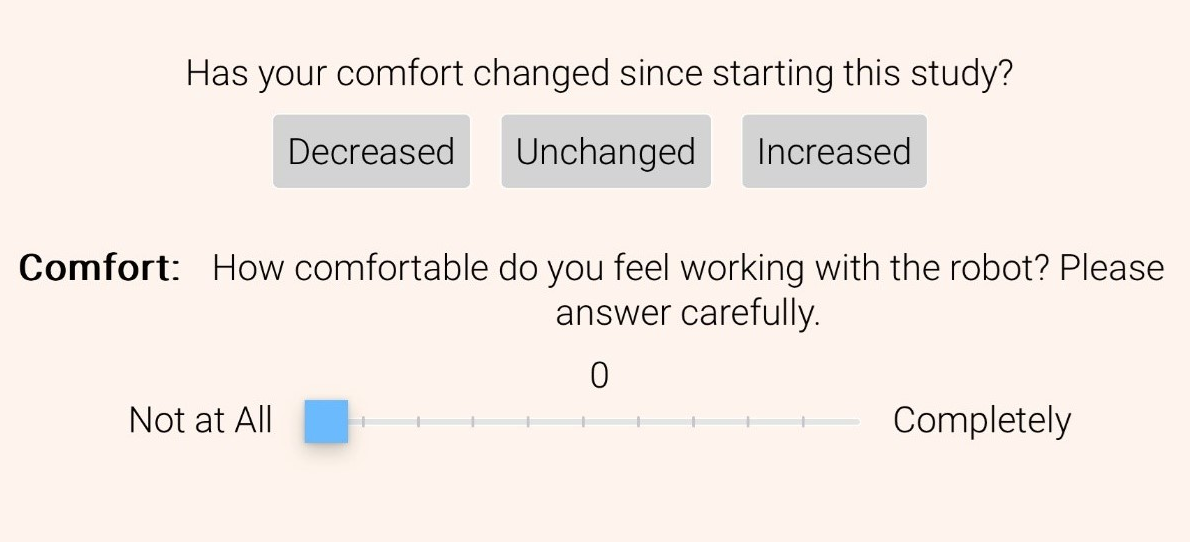}
    \caption{Our mixed-response questionnaire for comfort feedback is composed of a visual analog scale (continuous) and a relative comfort reporting question (discrete). Subjects complete the questionnaire after each episode.}
    \label{fig:vas-delta-scale}
\end{figure}

We developed a mixed-response format questionnaire for quick comfort feedback after every episode. The questionnaire is shown in Fig. \ref{fig:vas-delta-scale}. This mixed-response format questionnaire consists of two measures of comfort. 

The first measure of comfort is a \textbf{visual-analog scale (VAS)}~\cite{reipsIntervallevelMeasurementVisual2008} which captures users' subjective comfort on a continuous scale. Subjects see their comfort quantified between 0 and 10 at 0.1 intervals. We introduce metrics to measure the \textit{total comfort} and \textit{differential comfort} from the repeated mixed-response questionnaire, to capture zeroth-order and first-order dynamics of comfort, respectively. We define total comfort as the average comfort over interactions based on the VAS responses 
\begin{equation}\label{eq:total_comfort} 
        c_{total} = \frac{1}{K}\sum_{k=1}^{K}c(k)
\end{equation}
where $K$ is the number of interactions. We use averaged comfort since it is known to be better than final reports in quantifying user trust~\cite{yangEvaluatingEffectsUser2017a}.
We quantify differential comfort from the VAS as follows
\begin{equation}\label{eq:comfort_evolution} 
    c_{\Delta}(k) = c(k) - c(k-1),\ \forall k=2,\cdots,K
\end{equation}
Note that $c_{\Delta}(k)$ quantifies how comfort evolved as a function of episodes, and thus helps understand how novice users get habituated to robots through consecutive interactions.

We designed a second response format to assess the change in users' comfort as they experienced more episodes of interactions. Specifically, we designed a \textbf{multi-choice question} in which users can select one of three options to indicate how their comfort had evolved after each iteration: \textit{decreased}, \textit{unchanged}, or \textit{increased}. 

As a third way to measure comfort, we used \textbf{pre- and post-experiment questionnaires} based on the \emph{Robotic Social Attributes Scale (RoSAS)}~\cite{carpinellaRoboticSocialAttributes2017} Discomfort Likert scale. We use this mixed-response format to i) incentivize subjects to be consistent and thoughtful in their responses and ii) improve validity by using a verified subjective comfort measure supported by prior works.
\subsection{Objective Measures} \label{Objective Measures}
We report two objective measures inspired by commonly-used fluency measures~\cite{hoffmanEvaluatingFluencyHuman2019}. While fluency measures are typically used to assess the level of human-robot coordination, they also shed light on habituation dynamics. Specifically, we measure the time during which the user i) stalled their motion by remaining idle when they could have moved (U-STALL) and ii) concurrently moved with the robot (C-ACT). Both are computed as percentages of the total episode length.
We annotated a total of 452 videos (one video per episode per user) to infer U-STALL and C-ACT. We discarded 28 videos due to poor video quality. 
We directly labeled the frames during which the user paused during the task to compute U-STALL. 
To measure C-ACT, we subtracted U-STALL from the time duration between the start of the task and the end of the user's motion since the robot moved continuously throughout the episode. Two members of the study team annotated 50\% and 60\% of the data respectively, with 10\% overlap. Inter-rater reliability analysis showed significant agreement between raters (Cohen's $\kappa=0.86$, with a 1 second tolerance window).

\subsection{Subjects}
We recruited 40 participants from our university's student body. We only included participants who were physically capable of completing the task and importantly, had no prior experience working with or around physical robots at the undergraduate level or higher. We ensured the satisfaction of this selection criteria via a verbal confirmation, and questions in a pre-experiment questionnaire. The 40 participants were aged between 18 and 27 years (M=$22.8$, SD=$2.83$). Subjects were evenly distributed across the eight study conditions. All participants received \$10 USD as compensation.

\subsection{Procedure}
Before the experiment, subjects signed a consent form and filled out the pre-experiment questionnaire which recorded their demographics, prior robot experience, and initial comfort towards the robot before working with it. Subjects were able to see the idle physical robot prior to answering the pre-experiment questionnaire but were not able to get close to it or see it move. All subjects completed 12 episodes of the task (approximately 20 minutes) and answered questionnaires after every episode. After all 12 episodes, subjects answered the post-experiment questionnaire. The total duration of the study was approximately 30 minutes.

\section{Results and Discussion}
We first conducted a 3-way factorial ANOVA to test for significant differences in \textit{total comfort}  among the eight study conditions, as defined in Eq. (\ref{eq:total_comfort}).
\begin{figure}
    \centering
    \includegraphics[width=\linewidth]{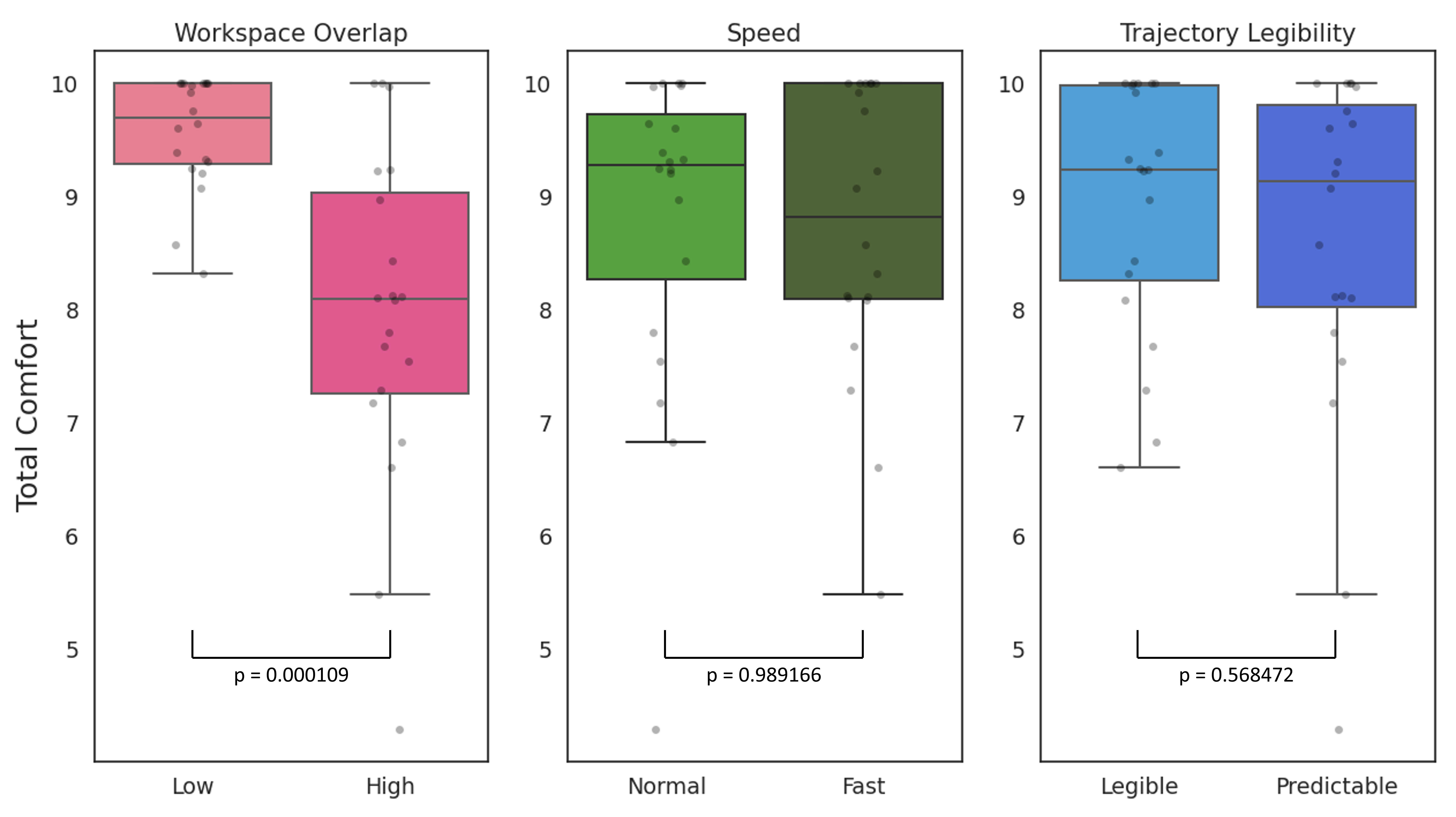}
    \caption{Total comfort with subject data partitioned by sub-conditions. Total comfort quantifies the subject's comfort after all interactions.}
    \label{fig:plot_vas_comfort_auc_hists}
\end{figure}
The 3-way factorial ANOVA revealed that only the workspace overlap condition had a statistically distinguished impact on total comfort (F(8, 32)=20.49, $p<0.01$). 
Tukey's HSD test found that the mean value of comfort was only significantly different between the \texttt{low} and \texttt{high} workspace overlap conditions ($p=0.01$). We found no significant individual or interaction effects from end-effector speed or trajectory legibility, nor from the interaction among all 3 conditions.

\subsection{Effects of motion characteristics on total comfort}\label{subsec:total_comfort_analyses}
We analyze the impact of the three motion characteristics on total comfort by partitioning the subject's data by the independent variable conditions. We used a Mann–Whitney U Test to test for statistical significance.
As seen in Fig. \ref{fig:plot_vas_comfort_auc_hists}, 
workspace overlap is the only significant factor ($p=0.0001$), with subjects reporting considerably higher comfort levels for lower workspace overlap. In contrast, speed ($p=0.99$) and legibility ($p=0.57$) do not have significant impacts on total comfort. These results suggest that low-overlap workspaces are desirable when introducing novice users to robots.

\begin{figure}
    \centering
    \includegraphics[width=\linewidth,trim={5 0 0 0}]{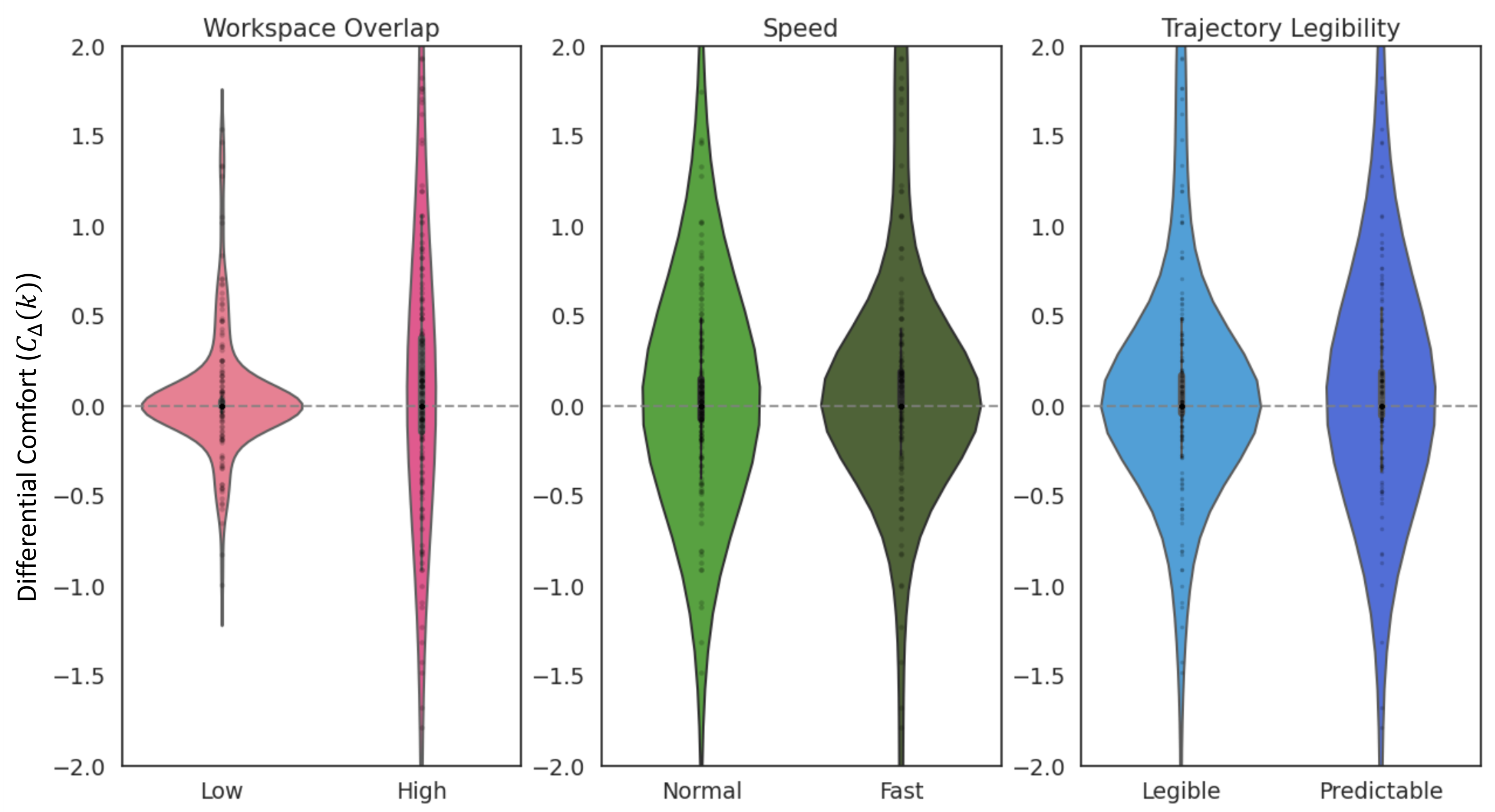}
    \caption{Violin plots of episodic differential comfort, partitioned by sub-conditions within each motion characteristic. Differential comfort computes the difference in comfort between consecutive episodes.}
    \label{fig:plot_vas_comfort_variation_violin}
\end{figure}
\subsection{Effects of motion characteristics on comfort dynamics}
\label{sec:Analysis of Comfort Variation for Workspace Overlap, Speed, and Legibility}

We next analyze the impact of the three motion characteristics on the subjects' differential comfort, based on VAS responses.
To this end, we show the distributions of $c_{\Delta}(k)$ values across each episode of each user as violin plots in Fig. \ref{fig:plot_vas_comfort_variation_violin}.
First, we note that all users were more likely to retain their comfort level between consecutive interactions rather than increase or decrease it. This is supported by the probability mass being centered around zero. This observation suggests that users gradually habituate to robots with relatively small updates to their comfort levels, irrespective of the robot's motion characteristics.
Second, we note that workspace overlap once again has the most pronounced impact on differential comfort, with low workspace overlap resulting in a considerably narrower distribution of $c_{\Delta}(k)$ than high workspace overlap. This suggests that low workspace overlap resulted in the most gradual and smooth habituation process for users, as evidenced by the thin-tailed distribution and fewer decreases in comfort between consecutive episodes.  

\begin{figure}
    \centering
    \includegraphics[width=\columnwidth]{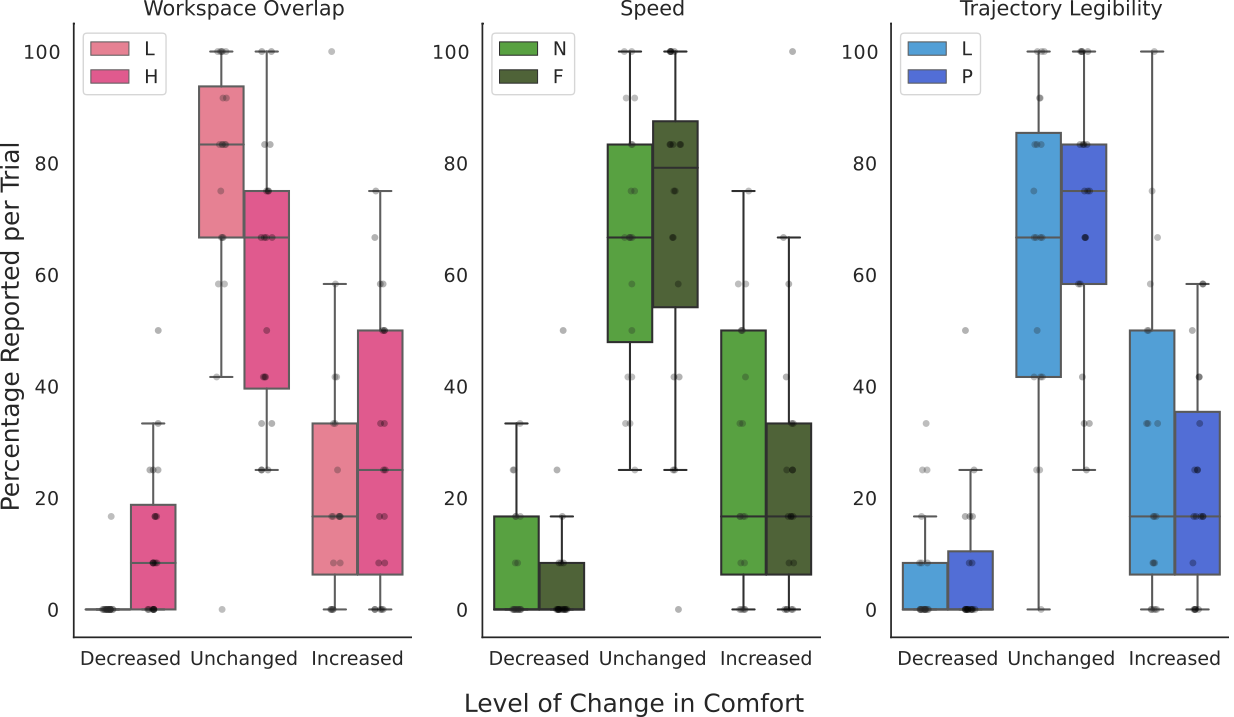}
    \caption{Users' comfort remained unchanged between two consecutive episodes more often than it increased or decreased. In contrast to speed and legibility, workspace overlap had the largest impact.}
    \label{fig:plot_rel_comfort_percentages_boxplots}
\end{figure}

We also analyzed users' responses to the multi-choice question. For each sub-condition of the motion characteristics, we show box plots in Fig. \ref{fig:plot_rel_comfort_percentages_boxplots} that indicate the percentage of episodes after which each user reported that their comfort \emph{decreased}, remained \emph{unchanged}, or \emph{increased}. Similar to our findings based on VAS, we observe that i) users are more likely to retain their sense of comfort between two consecutive episodes, and ii) workspace overlap is the only condition with a significant impact on users' perception of change in comfort. 
Notably, low workspace overlap results in considerably fewer instances of decrease in comfort, compared to high overlap.  

\begin{figure}[!ht]
    \centering
    \includegraphics[width=\linewidth,trim={0 0 0 4.7cm},clip]{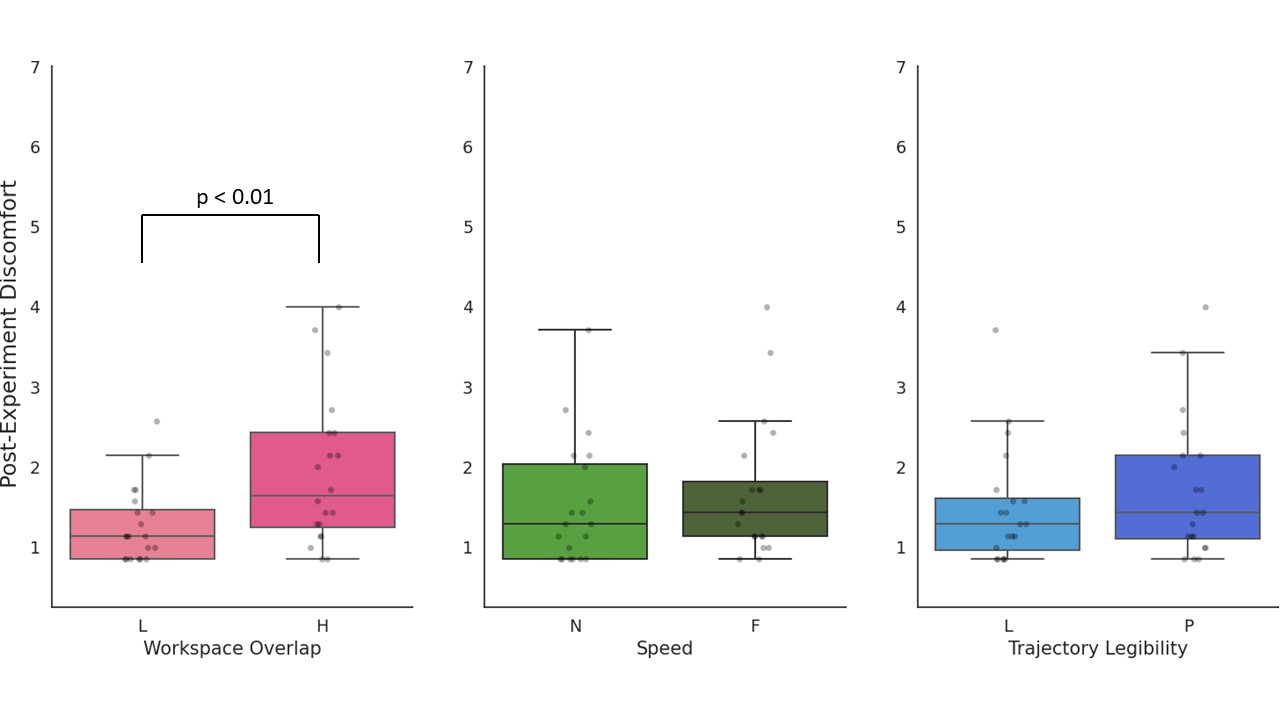}
    \caption{
    Post-experiment reports on the RoSAS discomfort scale (lower is better) suggest that workspace overlap most significantly impacts comfort.
    }
    \label{fig:plot_rosas_barplot_grouped_conditions}
\end{figure}

\subsection{Analysis of Pre- and Post-experiment Comfort} \label{sec:Analysis of Pre- and Post-experiment Comfort}
In Fig. \ref{fig:plot_rosas_barplot_grouped_conditions}, we report the post-experiment discomfort ratings from the RoSAS discomfort scale. Overall, subjective discomfort after 12 episodes is low for all motion IVs, suggesting that users indeed get habituated to working with the robot.  However, we observe that post-experiment discomfort is higher for high workspace overlap compared to all other motion conditions. We performed a Welch's t-test for each motion characteristic and found a significant difference between workspace overlap conditions ($p<0.01$). Speed and trajectory legibility did not have a significant impact on post-experiment discomfort. This result, when combined with 
the observations from our other analyses, strongly suggest that workspace overlap has a demonstrably larger impact on user comfort than does robot motion speed or legibility. 

\begin{figure}
    \centering
    \includegraphics[width=\linewidth]{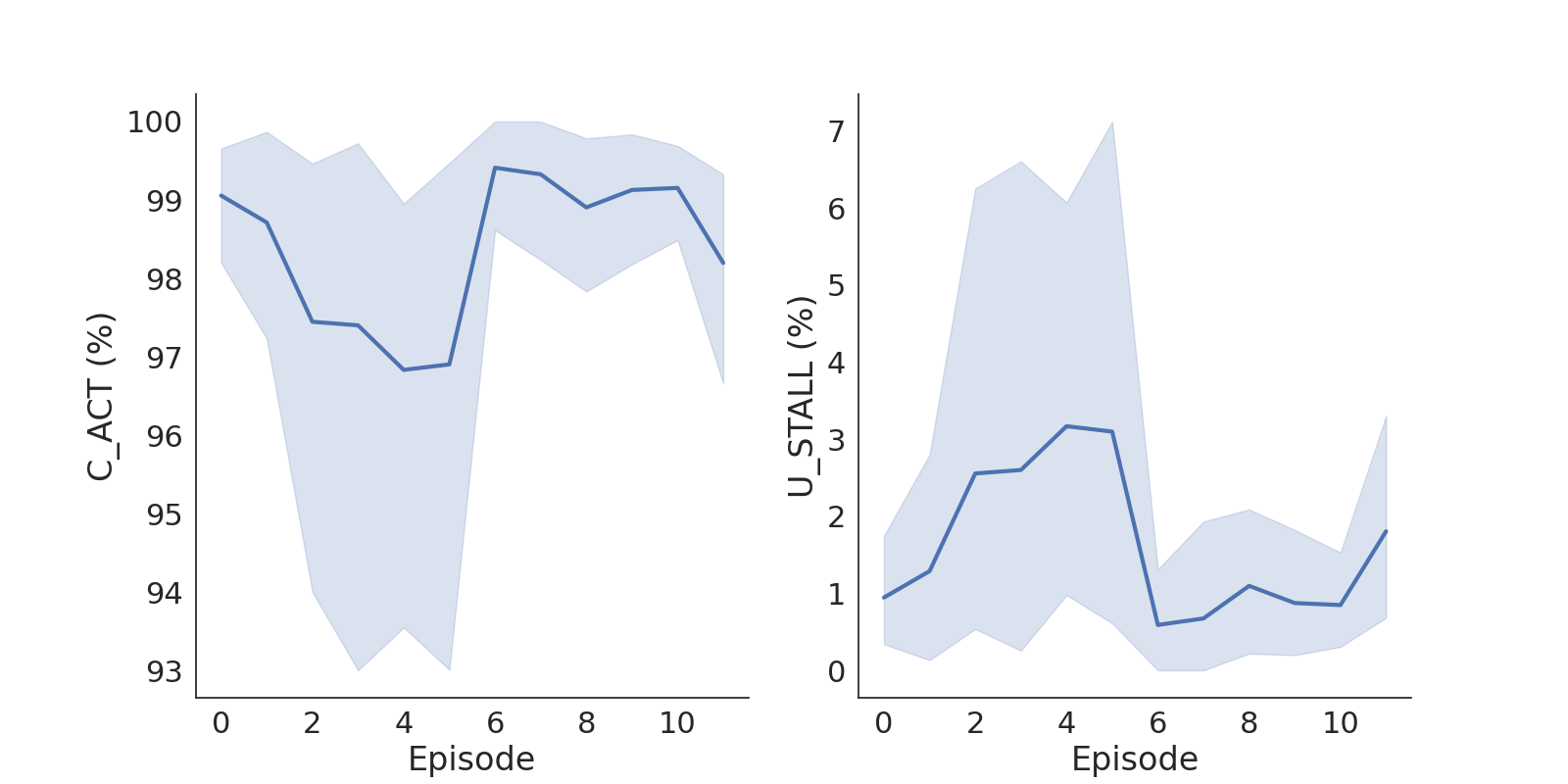}
    \caption{
    Novice users tend to move more concurrently with the robot (C-ACT) and stall less (U-STALL) as they gain more experience across different robot motion characteristics.
    }
    \label{fig:u_idle_all}
\end{figure}

\subsection{Effects of motion characteristics on objective measures} 
We show the mean and standard deviation of stalling time (U-STALL) of all users as a function of the episodes in Fig. \ref{fig:u_idle_all}. We observe that, in general, users tend to stall more during earlier episodes (2-5) and engage in more concurrent motion as they get more habituated to the robot. The stall time observed during later episodes is most likely due to users waiting for the robot to complete the sub-task of moving blocks. 

\begin{figure}
    \centering
    \includegraphics[width=0.7\linewidth,trim={50 10 70 10}]{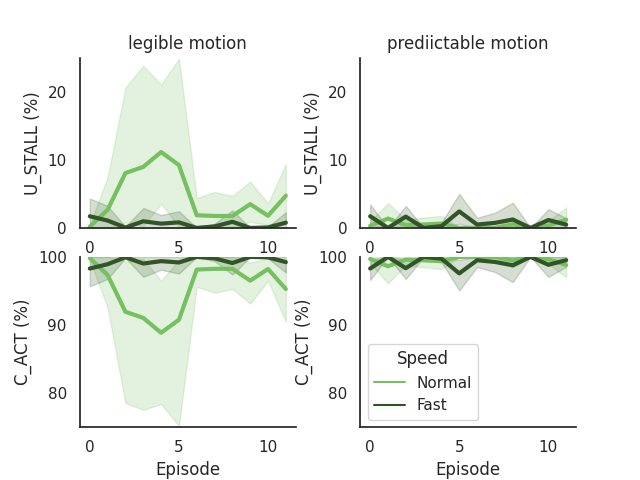}
    \caption{
    Robot speed had a considerable impact on concurrent movement and user stalling when the motion was legible (\textit{left}), but not when it was predictable (\textit{right}).
    }
    \label{fig:obj_speed_legibility}
\end{figure}

To analyze the effect of motion characteristics on fluency, we perform a 3-way factorial ANOVA test on U-STALL and C-ACT of all users for each episode. In stark contrast to comfort, we observe that only legibility and speed have statistically significant interaction effects on both of the metrics ($F(8, 32)=9.05, p<0.01$). To gain a deeper understanding of this interaction effect, we control for motion legibility and plot the evolution of U-STALL over 12 episodes (see Fig. \ref{fig:obj_speed_legibility}). 
Surprisingly, we find that users stall considerably more during initial episodes when the robot's motion is slow and legible compared to when it is fast and legible. We speculate that slow speed combined with legible trajectories may appear confusing to users, inducing them to pause their task to infer the robot's intention.
In contrast, we find that speed does not make a significant difference in how much users idle when robot motion is not legible. 

\section{Conclusion}

We study the effects of three robot motion characteristics -- workspace overlap, speed, and legibility -- on novice users' comfort and habituation over repeated interactions.
We found that workspace overlap, unlike speed and legibility, significantly impacts both total and differential comfort. Specifically, lower workspace overlap leads to significantly higher comfort levels and fewer fluctuations during habituation. We also found that users tended to stall less and move more concurrently with the robot after gaining experience, suggesting that habituation is possible in a relatively short period of time ($\sim$30 minutes). Unlike prior work, we did not find statistically significant interaction effects between motion characteristics. This difference could be explained by our experiment's relatively slow robot speed and the flexibility for users to move around the robot.

This work is a first step towards a comprehensive understanding of user comfort dynamics and the factors that influence it. However, our study could be improved in a number of ways. First, while our novice users do not have significant physical robot experience, all of them had a STEM background. Increasing the diversity and number of participants would more accurately capture the tendencies of the general population. Second, though we base our mixed-response questionnaire on prior work, it still needs to be validated. Third, more objective metrics can be employed to minimize reporting bias and strengthen our findings. Finally, the users never experienced robot task failures or sudden changes in motion characteristics; a potential line of future work is to evaluate user comfort habituation in the presence of robot failures or unexpected changes (e.g. sudden change in speed or proximity).

\addtolength{\textheight}{-12cm}   





\bibliographystyle{plain} 
\bibliography{refs} 

\end{document}


\section{Subjects}

\section{Supplementary Results}
In this section, we provide supplementary results for completeness and support of the primary results presented in the paper.

\noindent \textbf{Total Comfort Analysis of Variance}: The results from the 3-way factorial ANOVA for total comfort at the end of user's interaction is shown in Table \ref{table:total-comfort-anova-table}. 

\begin{table}[htbp]
\centering
\caption{3-Way Factorial ANOVA for \textit{Total Comfort}}
\begin{tabular}{|l|c|c|c|}
\hline
\textbf{Factor} & \textbf{F-Value} & \textbf{Sum of Squares} & \textbf{p-value} \\ \hline
Speed & 0.049722 & 0.063680 & 0.824965 \\ \hline

Workspace Overlap & 20.493743 & 26.246970 & 0.000078 \\ \hline

Legibility  & 0.895941 & 1.147459 & 0.350966 \\ \hline

Speed, Workspace Overlap & 0.043443 & 0.055639 & 0.836214 \\ \hline

Speed, Legibility & 0.950437 & 1.217254 & 0.336921 \\ \hline

Workspace Overlap, Legibility & 0.431138 &  0.552172 & 0.516126 \\ \hline

Speed, Workspace Overlap, Legibility & 1.338590 & 1.714374 & 0.255847 \\ \hline

\end{tabular}
\label{table:total-comfort-anova-table}
\end{table}

\noindent \textbf{Pre- and Post-Experiment Discomfort Difference}
In Fig. \ref{fig:rosas-pre-post-diff-barplot}, we report the difference in discomfort ratings from the RoSAS scale between pre- and post-experiment reports. Note the difference is pre-experiment discomfort minus post-experiment discomfort. Overall, high workspace overlap results in a greater variation of change in discomfort over the entire interaction.
\begin{figure}[htp]
    \centering
    \includegraphics[width=\textwidth]{figures/plot_rosas_pre_post_diff_barplot_grouped_conditions.png}
    \caption{Difference between pre- and post-experiment reports on the RoSAS 7-point discomfort scale (higher is better). These results suggest that workspace overlap most significantly impacts habituation comfort}
    \label{fig:rosas-pre-post-diff-barplot}
\end{figure}